\documentclass[11pt,a4paper]{article}
\usepackage[hyperref]{acl2018}
\usepackage{times}
\usepackage{latexsym}

\usepackage{bm}
\usepackage{graphicx}
\usepackage{amsmath}
\DeclareMathOperator*{\argmax}{arg\,max}

\usepackage{url}

\aclfinalcopy 

\usepackage{xcolor}

\newcommand{\mycomment}[3]{}

\newcommand{\ignore}[1]{}

\title{Bi-Directional Neural Machine Translation with Synthetic Parallel Data}

\author{Xing Niu \\
  University of Maryland \\
  {\tt \href{mailto:xingniu@cs.umd.edu}{xingniu@cs.umd.edu}} \\\And
  Michael Denkowski \\
  Amazon.com, Inc. \\
  {\tt \href{mailto:mdenkows@amazon.com}{mdenkows@amazon.com}} \\\And
  Marine Carpuat \\
  University of Maryland \\
  {\tt \href{mailto:marine@cs.umd.edu}{marine@cs.umd.edu}} \\}

\date{}

\begin{document}
\maketitle
\begin{abstract}
	Despite impressive progress in high-resource settings, Neural Machine Translation (NMT) still struggles in low-resource and out-of-domain scenarios, often failing to match the quality of phrase-based translation.
	We propose a novel technique that combines back-translation and multilingual NMT to improve performance in these difficult cases.
	Our technique trains a single model for both directions of a language pair, allowing us to back-translate source or target monolingual data without requiring an auxiliary model.
	We then continue training on the augmented parallel data, enabling a cycle of improvement for a single model that can incorporate any source, target, or parallel data to improve both translation directions.
	As a byproduct, these models can reduce training and deployment costs significantly compared to uni-directional models.
	Extensive experiments show that our technique outperforms standard back-translation in low-resource scenarios, improves quality on cross-domain tasks, and effectively reduces costs across the board.
\end{abstract}

\section{Introduction}

Neural Machine Translation (NMT) has been rapidly adopted in industry as it consistently outperforms previous methods across domains and language pairs \citep{BojarCFGHHHKLLM17,IWSLT2017}.
However, NMT systems still struggle compared to Phrase-based Statistical Machine Translation (SMT) in low-resource or out-of-domain scenarios \citep{KoehnK17}.
This performance gap is a significant roadblock to full adoption of NMT.

In many low-resource scenarios, parallel data is prohibitively expensive or otherwise impractical to collect, whereas monolingual data may be more abundant.
SMT systems have the advantage of a dedicated language model that can incorporate all available target-side monolingual data to significantly improve translation quality \citep{KoehnOM03,KoehnS07}.
By contrast, NMT systems consist of one large neural network that performs full sequence-to-sequence translation \citep{SutskeverVL14,ChoMGBBSB14}.
Trained end-to-end on parallel data, these models lack a direct avenue for incorporating monolingual data.
\citet{SennrichHB16} overcome this challenge by back-translating target monolingual data to produce \textit{synthetic} parallel data that can be added to the training pool.
While effective, back-translation introduces the significant cost of first building a reverse system.

Another technique for overcoming a lack of data is multitask learning, in which domain knowledge can be transferred between related tasks \citep{Caruana97}.
\citet{JohnsonSLKWCTVW17} apply the idea to multilingual NMT by concatenating parallel data of various language pairs and marking the source with the desired output language.
The authors report promising results for translation between languages that have zero parallel data. This approach also dramatically reduces the complexity of deployment by packing multiple language pairs into a single model.

We propose a novel combination of back-translation and multilingual NMT that trains both directions of a language pair jointly in a single model.
Specifically, we initialize a bi-directional model on parallel data and then use it to translate select source and target monolingual data.
Training is then continued on the augmented parallel data, leading to a cycle of improvement.
This approach has several advantages:

\begin{itemize}
	\item A single NMT model with standard architecture that performs all forward and backward translation during training.
	\item Training costs reduced significantly compared to uni-directional systems.
	\item Improvements in translating quality for low-resource languages, even over uni-directional systems with back-translation.
	\item Effectiveness in domain adaptation.
\end{itemize}

Via comprehensive experiments, we also contribute to best practices in selecting most suitable combinations of synthetic parallel data and choosing appropriate amount of monolingual data.

\section{Approach}

In this section, we introduce an efficient method for improving bi-directional neural machine translation with synthetic parallel data.
We also present a strategy for selecting suitable monolingual data for back-translation.

\subsection{Bi-Directional NMT with Synthetic Parallel Data}

We use the techniques described by \citet{JohnsonSLKWCTVW17} to build a multilingual model that combines forward and backward directions of a single language pair.
To begin, we construct training data by swapping the source and target sentences of a parallel corpus and appending the swapped version to the original.
We then add an artificial token to the beginning of each source sentence to mark the desired target language, such as $<$2en$>$ for English.
A standard NMT system can then be trained on the augmented dataset, which is naturally balanced between language directions.\footnote{\citet{JohnsonSLKWCTVW17} report the need to oversample when data is significantly unbalanced between language pairs.}
A shared Byte-Pair Encoding (BPE) model is built on source and target data, alleviating the issue of unknown words and reducing the vocabulary to a smaller set of items shared across languages \citep{SennrichHB16a,JohnsonSLKWCTVW17}.
We further reduce model complexity by tying source and target word embeddings.
The full training process significantly saves the total computing resources compared to training an individual model for each language direction.

Generating synthetic parallel data is straightforward with a bi-directional model: sentences from both source and target monolingual data can be translated to produce synthetic sentence pairs.
Synthetic parallel data of the form \texttt{synthetic $\rightarrow$ monolingual} can then be used in the forward direction, the backward direction, or both.
Crucially, this approach leverages both source and target monolingual data while always placing the real data on the target side, eliminating the need for work-arounds such as freezing certain model parameters to avoid degradation from training on MT output \citep{ZhangZ16}.

\subsection{Monolingual Data Selection}
\label{sec:data-selection}

Given the goal of improving a base bi-directional model, selecting ideal monolingual data for back-translation presents a significant challenge.
Data too close to the original training data may not provide sufficient new information for the model.
Conversely, data too far from the original data may be translated too poorly by the base model to be useful.
We manage these risks by leveraging a standard pseudo in-domain data selection technique, cross-entropy difference \citep{MooreL10,AxelrodHG11}, to rank sentences from a general domain.
Smaller cross-entropy difference indicates a sentence that is simultaneously more similar to the in-domain corpus (e.g. real parallel data) and less similar to the average of the general-domain monolingual corpus.
This allows us to begin with ``safe'' monolingual data and incrementally expand to higher risk but potentially more informative data.




\section{Experiments}
\label{sec:exp}

In this section, we describe data, settings, and experimental methodology.
We then present the results of comprehensive experiments designed to answer the following questions:
(1) How can synthetic data be most effectively used to improve translation quality?
(2) Does the reduction in training time for bi-directional NMT come at the cost of lower translation quality?
(3) Can we further improve training speed and translation quality training with incremental training and re-decoding?
(4) How can we effectively choose monolingual training data?
(5) How well does bi-directional NMT perform on domain adaptation?

\subsection{Data}

\begin{table}[t]
	\begin{center}
		\begin{tabular}{l|l|r}
			Type & Dataset & \# Sentences\\
			\hline
			\multicolumn{3}{l}{High-resource: German$\leftrightarrow$English} \\
			\hline
			Training & Common Crawl + & \\
			& Europarl v7 + & \\
			& News Comm. v12 & 4,356,324 \\
			Dev & Newstest 2015+2016 & 5,168 \\
			Test & Newstest 2017 & 3,004 \\
			Mono-DE & News Crawl 2016 & 26,982,051 \\
			Mono-EN & News Crawl 2016 & 18,238,848 \\
			\hline
			\multicolumn{3}{l}{Low-resource: Tagalog$\leftrightarrow$English} \\
			\hline
			Training & News/Blog & 50,705 \\
			Dev/Test & News/Blog & 491/508 \\
			Dev/Test* & Bible & 500/500 \\
			Sample* & Bible & 61,195 \\
			Mono-TL & Common Crawl & 26,788,048 \\
			Mono-EN & ICWSM 2009 blog & 48,219,743 \\
			\hline
			\multicolumn{3}{l}{Low-resource: Swahili$\leftrightarrow$English} \\
			\hline
			Training & News/Blog & 23,900 \\
			Dev/Test & News/Blog & 491/509 \\
			Dev/Test* & Bible-NT & 500/500 \\
			Sample* & Bible-NT & 14,699 \\
			Mono-SW & Common Crawl & 12,158,524 \\
			Mono-EN & ICWSM 2009 blog & 48,219,743 \\
			\hline
		\end{tabular}
	\end{center}
	\caption{Data sizes of training, development, test, sample and monolingual sets. Sample data serves as the in-domain seed for data selection.}
	\label{tab:data}
\end{table}


\paragraph{Diverse Language Pairs:}
We evaluate our approach on both high and low-resource data sets: German$\leftrightarrow$English (\texttt{DE$\leftrightarrow$EN}), Tagalog$\leftrightarrow$English \texttt{TL$\leftrightarrow$EN}, and Swahili$\leftrightarrow$English (\texttt{SW$\leftrightarrow$EN}).
Parallel and monolingual \texttt{DE$\leftrightarrow$EN} data are provided by the WMT17 news translation task \citep{BojarCFGHHHKLLM17}.
Parallel data for \texttt{TL$\leftrightarrow$EN} and \texttt{SW$\leftrightarrow$EN} contains a mixture of domains such as news and weblogs, and is provided as part of the IARPA MATERIAL program.\footnote{\url{https://www.iarpa.gov/index.php/research-programs/material}} We split the original corpora into training, dev, and test sets, therefore they share a homogeneous n-gram distribution.
For these low-resource pairs, \texttt{TL} and \texttt{SW} monolingual data are provided by the Common Crawl \citep{BuckHO14} while \texttt{EN} monolingual data is provided by the ICWSM 2009 Spinn3r blog dataset (tier-1) \citep{BurtonJS09}.

\paragraph{Diverse Domain Settings:}
For WMT17 \texttt{DE$\leftrightarrow$EN}, we choose news articles from 2016 (the closest year to the test set) as in-domain data for back-translation.
For \texttt{TL$\leftrightarrow$EN} and \texttt{SW$\leftrightarrow$EN}, we identify in-domain and out-of-domain monolingual data and apply data selection to choose pseudo in-domain data (see Section~\ref{sec:data-selection}).
We use the training data as in-domain and either Common Crawl or ICWSM as out-of-domain.
We also include a low-resource, long-distance domain adaptation task for these languages: training on News/Blog data and testing on Bible data.
We split a parallel Bible corpus \citep{Christodoulopoulos15} into sample, dev, and test sets, using the sample data as the in-domain seed for data selection.

\paragraph{Preprocessing:}
Following \citet{CoRR:Sockeye}, we apply four pre-processing steps to parallel data: normalization, tokenization, sentence-filtering (length 80 cutoff), and joint source-target BPE with 50,000 operations \citep{SennrichHB16a}.
Low-resource language pairs are also true-cased to reduce sparsity. BPE and true-casing models are rebuilt whenever the training data changes.
Monolingual data for low-resource settings is filtered by retaining sentences longer than nine tokens.
Itemized data statistics after preprocessing can be found in Table~\ref{tab:data}.

\begin{table*}[t]
	\begin{center}
		\scalebox{0.93}{
			\begin{tabular}{l|l|rr|rr|rr}
				ID & Training Data & \texttt{TL}$\rightarrow$\texttt{EN} & \texttt{EN}$\rightarrow$\texttt{TL} & \texttt{SW}$\rightarrow$\texttt{EN} & \texttt{EN}$\rightarrow$\texttt{SW} & \texttt{DE}$\rightarrow$\texttt{EN} & \texttt{EN}$\rightarrow$\texttt{DE} \\
				\hline
				U-1 & \texttt{L1}$\rightarrow$\texttt{L2} & 31.99 & 31.28 & 32.60 & 39.98 & 29.51 & 23.01 \\
				\cline{3-6}
				U-2 & \texttt{L1}$\rightarrow$\texttt{L2} + \texttt{L1*}$\rightarrow$\texttt{L2} & \bf 24.21 & \bf 29.68 & \bf 25.84 & \bf 38.29 & \bf 33.20 & \bf 25.41 \\
				U-3 & \texttt{L1}$\rightarrow$\texttt{L2} \hspace{53pt} + \texttt{L1}$\rightarrow$\texttt{L2*} & 22.13 & 27.14 & 24.89 & 36.53 & 30.89 & 23.72 \\
				U-4 & \texttt{L1}$\rightarrow$\texttt{L2} + \texttt{L1*}$\rightarrow$\texttt{L2} + \texttt{L1}$\rightarrow$\texttt{L2*} & 23.38 & 29.31 & 25.33 & 37.46 & 33.01 & 25.05 \\
				\hline
				\hline
				& \multicolumn{1}{c}{\texttt{L1}$=$\texttt{EN}} & \multicolumn{2}{|c}{\texttt{L2}$=$\texttt{TL}} & \multicolumn{2}{|c}{\texttt{L2}$=$\texttt{SW}} & \multicolumn{2}{|c}{\texttt{L2}$=$\texttt{DE}} \\
				\hline
				B-1 & \texttt{L1}$\leftrightarrow$\texttt{L2} & 32.72 & 31.66 & 33.59 & 39.12 & 28.84 & 22.45 \\
				B-2 & \texttt{L1}$\leftrightarrow$\texttt{L2} + \texttt{L1*}$\leftrightarrow$\texttt{L2} & 32.90 & \bf 32.33 & 33.70 & \bf 39.68 & 29.17 & \bf 24.45 \\
				B-3 & \texttt{L1}$\leftrightarrow$\texttt{L2} \hspace{53pt} + \texttt{L2*}$\leftrightarrow$\texttt{L1} & 32.71 & 31.10 & 33.70 & 39.17 & \bf 31.71 & 21.71 \\
				B-4 & \texttt{L1}$\leftrightarrow$\texttt{L2} + \texttt{L1*}$\leftrightarrow$\texttt{L2}  + \texttt{L2*}$\leftrightarrow$\texttt{L1} & \bf 33.25 & \bf 32.46 & \bf 34.23 & 38.97 & 30.43 & 22.54 \\
				B-5 & \texttt{L1}$\leftrightarrow$\texttt{L2} + \texttt{L1*}$\rightarrow$\texttt{L2}  + \texttt{L2*}$\rightarrow$\texttt{L1} & \bf 33.41 & \bf 33.21 & \bf 34.11 & \bf 40.24 & \bf 31.83 & \bf 24.61 \\
				\hline
				B-5* & \texttt{L1}$\leftrightarrow$\texttt{L2} + \texttt{L1*}$\rightarrow$\texttt{L2}  + \texttt{L2*}$\rightarrow$\texttt{L1} & 33.79 & 32.97 & 34.15 & 40.61 & 31.94 & 24.45 \\
				B-6* & \texttt{L1}$\leftrightarrow$\texttt{L2} + \underline{\texttt{L1*}}$\rightarrow$\texttt{L2}  + \underline{\texttt{L2*}}$\rightarrow$\texttt{L1} & \bf 34.50 & \bf 33.73 & \bf 34.88 & \bf 41.53 & \bf 32.49 & \bf 25.20 \\
				\hline
		\end{tabular}}
	\end{center}
	\caption{BLEU scores for uni-directional models (U-*) and bi-directional NMT models (B-*) trained on different combinations of real and synthetic parallel data. Models in B-5* are fine-tuned from base models in B-1. Best models in B-6* are fine-tuned from precedent models in B-5* and underscored synthetic data is re-decoded using precedent models. Scores with largest improvement within each zone are highlighted.}
	\label{tab:nmt}
\end{table*}

\subsection{NMT Configuration}

We use the attentional RNN encoder-decoder architecture implemented in the Sockeye toolkit \citep{CoRR:Sockeye}.
Our translation model uses a bi-directional encoder with a single LSTM layer of size 512, multilayer perceptron attention with a layer size of 512, and word representations of size 512 \citep{BahdanauCB15}.
We apply layer normalization \citep{BaKH16} and tie source and target embedding parameters.
We train using the Adam optimizer with a batch size of 64 sentences and checkpoint the model every 1000 updates (10,000 for \texttt{DE$\leftrightarrow$EN}) \citep{KingmaB15}.
Training stops after 8 checkpoints without improvement of perplexity on the development set.
We decode with a beam size of 5.

For \texttt{TL$\leftrightarrow$EN} and \texttt{SW$\leftrightarrow$EN}, we add dropout to embeddings and RNNs of the encoder and decoder with probability 0.2.
We also tie the output layer's weight matrix with the source and target embeddings to reduce model size \citep{PressW17}.
The effectiveness of tying input/output target embeddings has been verified on several low-resource language pairs \citep{NguyenC18}.

For \texttt{TL$\leftrightarrow$EN} and \texttt{SW$\leftrightarrow$EN}, we train four randomly seeded models for each experiment and combine them in a linear ensemble for decoding.
For \texttt{DE$\leftrightarrow$EN} experiments, we train a single model and average the parameters of the best four checkpoints for decoding \citep{Junczys-Dowmunt16}.
We report case-insensitive BLEU with standard WMT tokenization.\footnote{We use the script \url{https://github.com/EdinburghNLP/nematus/blob/master/data/multi-bleu-detok.perl}}

\subsection{Uni-Directional NMT}

We first evaluate the impact of synthetic parallel data on standard uni-directional NMT.
Baseline systems trained on real parallel data are shown in row U-1 of Table~\ref{tab:nmt}.\footnote{Baseline BLEU scores are higher than expected on low-resource language pairs. We hypothesize that the data is homogeneous and easier to translate.}
In all tables, we use \texttt{L1}$\rightarrow$\texttt{L2} to indicate real parallel data where the source language is \texttt{L1} and the target language is \texttt{L2}.
Synthetic data is annotated by asterisks, such as \texttt{L1*}$\rightarrow$\texttt{L2} indicating that \texttt{L1*} is the synthetic back-translation of real monolingual data \texttt{L2}.

We always select monolingual data as an integer multiple of the amount of real parallel data $n$, i.e. $|$\texttt{L1}$\rightarrow$\texttt{L2*}$|$ = $|$\texttt{L1*}$\rightarrow$\texttt{L2}$|$ = $kn$.
For \texttt{DE$\leftrightarrow$EN} models, we simply choose the top-$n$ sentences from shuffled News Crawl corpus.
For all models of low-resource languages, we select the top-$3n$ sentences ranked by cross-entropy difference as described in Section~\ref{sec:data-selection}. The choice of $k$ is discussed in Section~\ref{sec:mono-size}.

Shown in rows U-2 through U-4 of Table~\ref{tab:nmt}, we compare the results of incorporating different combinations of real and synthetic parallel data.
Models trained on \textbf{only real data of target language} (i.e. in U-2) achieve better performance in BLEU than using other combinations.
This is an expected result since translation quality is highly correlated with target language models.
By contrast, standard back-translation is not effective for our low-resource scenarios.
A significant drop ($\sim$7 BLEU comparing U-1 and U-2 for \texttt{TL/SW}$\rightarrow$\texttt{EN}) is observed when back-translating English.
One possible reason is that the quality of the selected monolingual data, especially English, is not ideal. We will encounter this issue again when using bi-directional models with the same data in Section~\ref{sec:bi-nmt}.

\subsection{Bi-Directional NMT}
\label{sec:bi-nmt}

\begin{table*}[t]
	\begin{center}
		\scalebox{1}{
			\begin{tabular}{c|l|r|r|r|r|r|r}
				\multicolumn{2}{c|}{Model} & \texttt{TL}$\rightarrow$\texttt{EN} & \texttt{EN}$\rightarrow$\texttt{TL} & \texttt{SW}$\rightarrow$\texttt{EN} & \texttt{EN}$\rightarrow$\texttt{SW} & \texttt{DE}$\rightarrow$\texttt{EN} & \texttt{EN}$\rightarrow$\texttt{DE} \\
				\hline
				& Baseline & 76 & 78 & 63 & 66 & 41 & 48 \\
				Uni-directional & Synthetic & 177 & 176 & 137 & 104 & 88 & 75 \\
				& TOTAL & \multicolumn{2}{|r}{507} & \multicolumn{2}{|r}{371} & \multicolumn{2}{|r}{252} \\
				\hline
				& Baseline & \multicolumn{2}{|r}{125} & \multicolumn{2}{|r}{93} & \multicolumn{2}{|r}{61} \\
				Bi-directional & Synthetic & \multicolumn{2}{|r}{285} & \multicolumn{2}{|r}{218} & \multicolumn{2}{|r}{113} \\
				& TOTAL & \multicolumn{2}{|r}{$\downarrow$ 19\%\hspace{30pt}410} & \multicolumn{2}{|r}{$\downarrow$ 14\%\hspace{30pt}311} & \multicolumn{2}{|r}{$\downarrow$ 31\%\hspace{30pt}174} \\
				\cline{2-8}
				(fine-tuning) & Synthetic & \multicolumn{2}{|r}{$\downarrow$ 23\%\hspace{30pt}219} & \multicolumn{2}{|r}{$\downarrow$ 44\%\hspace{30pt}122} & \multicolumn{2}{|r}{$\downarrow$ 24\%\hspace{35pt}86} \\
				\hline
		\end{tabular}}
	\end{center}
	\caption{Number of checkpoints ($=|\texttt{updates}|/$1000 for \texttt{TL/SW$\leftrightarrow$EN} or $|\texttt{updates}|/$10,000 for \texttt{DE$\leftrightarrow$EN}) used by various NMT models. Bi-directional models reduce the training time by 15-30\% (comparing `TOTAL' rows). Fine-tuning bi-directional baseline models on synthetic parallel data reduces the training time by 20-40\% (comparing `Synthetic' rows).}
	\label{tab:bi-nmt-cp}
\end{table*}

We map the same synthetic data combinations to bi-directional NMT, comparing against uni-directional models with respect to both translation quality and training time.
Training bi-directional models requires doubling the training data by adding a second copy of the parallel corpus where the source and target are swapped.
We use the notation \texttt{L1}$\leftrightarrow$\texttt{L2} to represent the concatenation of \texttt{L1}$\rightarrow$\texttt{L2} and its swapped copy \texttt{L2}$\rightarrow$\texttt{L1} in Table~\ref{tab:nmt}.

Compared to independent models (i.e. U-1), the bi-directional \texttt{DE$\leftrightarrow$EN} model in B-1 is slightly worse (by $\sim$0.6~BLEU).
These losses match observations by \citet{JohnsonSLKWCTVW17} on many-to-many multilingual NMT models.
By contrast, bi-directional low-resource models slightly outperform independent models.
We hypothesize that in low-resource scenarios the neural model's capacity is far from exhausted due to the redundancy in neural network parameters \citep{DenilSDRF13}, and the benefit of training on twice as much data surpasses the detriment of confusing the model by mixing two languages.

We generate synthetic parallel data from the same monolingual data as in the uni-directional experiments.
If we build training data symmetrically (i.e. B-2,3,4), back-translated sentences are distributed equally on the source and target sides, forcing the model to train on some amount of synthetic target data (MT output).
For \texttt{DE$\leftrightarrow$EN} models, the best BLEU scores are achieved when synthetic training data is only present on the source side while for low-resource models, the results are mixed.
We see a particularly counter-intuitive result when using monolingual English data --- no significant improvement (see B-3 for \texttt{TL/SW}$\rightarrow$\texttt{EN}).
As bi-directional models are able to leverage monolingual data of both languages, better results are achieved when combining all synthetic parallel data (see B-4 for \texttt{TL/SW}$\rightarrow$\texttt{EN}).
By further excluding potentially harmful target-side synthetic data (i.e. B-4 $\rightarrow$ B-5), the most unified and slim models achieve best overall performance.

While the best bi-directional NMT models thus far (B-5) outperform the best uni-directional models (U-1,2) for low-resource language pairs, they struggle to match performance in the high-resource \texttt{DE$\leftrightarrow$EN} scenario.

In terms of efficiency, bi-directional models consistently reduce the training time by 15-30\% as shown in Table~\ref{tab:bi-nmt-cp}. Note that checkpoints are summed over all independent runs when ensemble decoding is used.

\subsubsection{Fine-Tuning and Re-Decoding}

Training new NMT models from scratch after generating synthetic data is incredibly expensive, working against our goal of reducing the overall cost of deploying strong translation systems.
Following the practice of mixed fine-tuning proposed by \citet{ChuDK17}, we continue training baseline models on augmented data as shown in B-5* of Table~\ref{tab:nmt}.
These models achieve comparable translation quality to those trained from scratch (B-5) at a significantly reduced cost, up to 20-40\% computing time in the experiments illustrated in Table~\ref{tab:bi-nmt-cp}.

We also explore re-decoding the same monolingual data using improved models \citep{SennrichHB16}.
Underscored synthetic data in B-6* is re-decoded by models in B-5*, leading to the best results for all low-resource scenarios and competitive results for our high-resource scenario.

%
%
%

\subsubsection{Size of Selected Monolingual Data}
\label{sec:mono-size}

\begin{figure*}[t]
	\centering
	\includegraphics[width=0.7\textwidth]{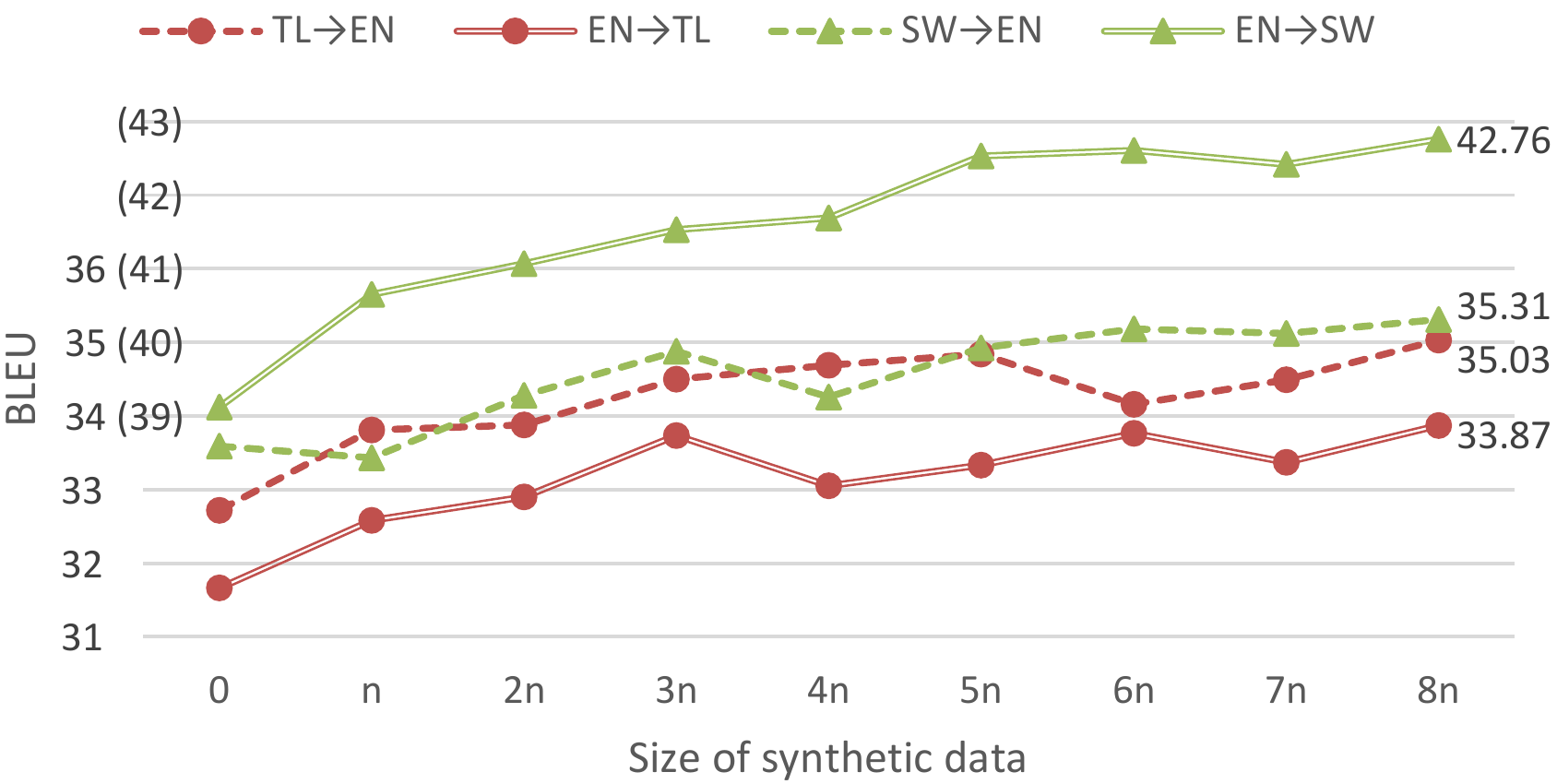}
	\caption{BLEU scores for four translation directions vs. the size of selected monolingual data. $n$ in x-axis equals to the size of real parallel data. \texttt{EN}$\rightarrow$\texttt{SW} models use BLEU in parentheses in y-axis. All language pairs have an increasing momentum and tend to converge with more synthetic parallel data.}
	\label{fig:mono-size}
\end{figure*}

In our experiments, the optimal amount of monolingual data for constructing synthetic parallel data is task-dependent.
Factors such as size and linguistic distribution of data and overlap between real parallel data, monolingual data, and test data can influence the effectiveness curve of synthetic data.
We illustrate the impact of varying the size of selected monolingual data in our low-resource scenario.
Shown in Figure~\ref{fig:mono-size}, all language pairs have an increasing momentum and tend to converge with more synthetic parallel data.
The optimal point is a hyper-parameter that can be empirically determined.

\subsubsection{Domain Adaptation}

\begin{table*}[t]
	\begin{center}
		\scalebox{1}{
			\begin{tabular}{l|l|rr|rr}
				\multicolumn{2}{}{} & \multicolumn{2}{|c}{\texttt{L2}$=$\texttt{TL}} & \multicolumn{2}{|c}{\texttt{L2}$=$\texttt{SW}} \\
				\cline{3-6}
				ID & Training Data (\texttt{L1}$=$\texttt{EN}) & \texttt{TL}$\rightarrow$\texttt{EN} & \texttt{EN}$\rightarrow$\texttt{TL} & \texttt{SW}$\rightarrow$\texttt{EN} & \texttt{EN}$\rightarrow$\texttt{SW} \\
				\hline
				A-1 & \texttt{L1}$\leftrightarrow$\texttt{L2} & 11.03 & 10.17 & 6.56 & 3.80 \\
				A-2 & \texttt{L1}$\leftrightarrow$\texttt{L2} + \texttt{L1*}$\rightarrow$\texttt{L2}  + \texttt{L2*}$\rightarrow$\texttt{L1} & 16.49 & 22.33 & 8.70 & 7.47 \\
				A-3 & \texttt{L1}$\leftrightarrow$\texttt{L2} + \underline{\texttt{L1*}}$\rightarrow$\texttt{L2}  + \underline{\texttt{L2*}}$\rightarrow$\texttt{L1} & \bf 18.91 & \bf 23.41 & \bf 11.01 & \bf 8.06 \\
				\hline
		\end{tabular}}
	\end{center}
	\caption{BLEU scores for bi-directional NMT models on Bible data. Models in A-2 are fine-tuned from baseline models in A-1. Highlighted best models in A-3 are fine-tuned from precedent models in A-2 and underscored synthetic data is re-decoded using precedent models. Baseline models are significantly improved in terms of BLEU.}
	\label{tab:domain-adapt}
\end{table*}

We evaluate the performance of using the same bi-directional NMT framework on a long-distance domain adaptation task: News/Blog to Bible.
This task is particularly challenging because out-of-vocabulary rates of Bible test sets are as high as 30-45\% when training on News/Blog.
Significant linguistic differences also exist between modern and Biblical language use.
The impact of this domain mismatch is demonstrated by the incredibly low BLEU scores of baseline News/Blog systems (Table~\ref{tab:domain-adapt}, A-1).
After fine-tuning baseline models on augmented parallel data (A-2) and re-decoding (A-3),\footnote{The concatenation of development sets from both News/Blog and Bible serves for validation.} we see BLEU scores increase by 70-130\%. Despite being based on extremely weak baseline performance, they still show the promise of our approach for domain adaptation.

\section{Related Work}

Leveraging monolingual data in NMT is challenging. For example, integrating language models in the decoder \citep{GulcehreFXCBLBS15} or initializing the encoder and decoder with pre-trained language models \citep{RamachandranLL17} would require significant changes to system architecture.

In this work, we build on the elegant and effective approach of turning incomplete (monolingual) data into complete (parallel) data by back-translation. \citet{SennrichHB16} used an auxiliary reverse-directional NMT system to generate synthetic source data from real monolingual target data, with promising results (+3 BLEU on strong baselines). Symmetrically, \citet{ZhangZ16} used an auxiliary same-directional translation system to generate synthetic target data from the real source language. However, parameters of the decoder have to be frozen while training on synthetic data, otherwise the decoder would fit to noisy MT output. By contrast, our approach effectively leverages synthetic data from both translation directions, with consistent gains in translation quality. A similar idea is used by \citet{ZhangLLZC18} with a focus on re-decoding iteratively. However, their NMT models of both directions are still trained independently.

Another technique for using monolingual data in NMT is round-trip machine translation. Suppose sentence $f$ from a monolingual dataset is translated forward to $e$ and then translated back to $f'$, then $f'$ and $f$ should be identical \citep{Brislin70}. \citet{ChengXHHWSL16} optimize $\argmax_{\theta} P(f'|f;\theta)$ as an autoencoder; \citet{WangXZBQLL18} minimize the difference between $P(f)$ and $P(f'|\theta)$ based on the law of total probability, while \citet{HeXQWYLM16} set the quality of both $e$ and $f'$ as rewards for reinforcement learning. They all achieve promising improvement but rely on non-standard training frameworks.

\ignore{One typical approach to integrate domain knowledge from various data is multitask (multilingual) NMT: models to be trained on different parallel datasets can be combined by sharing certain components.}
Multitask learning has been used in past work to combine models trained on different parallel corpora by sharing certain components. These components, such as the attention mechanism \citep{FiratCB16}, benefit from being trained on an effectively larger dataset. In addition, the more parameters are shared, the faster a joint model can be trained --- this is particularity beneficial in industry settings. Baidu built one-to-many translation systems by sharing both encoder and attention \citep{DongWHYW15}. Google enabled a standard NMT framework to support many-to-many translation directions by simply attaching a language specifier to each source sentence \citep{JohnsonSLKWCTVW17}. We adopted Google's approach to build bi-directional systems that successfully combine actual and synthetic parallel data.

\section{Conclusion}

We propose a novel technique for bi-directional neural machine translation.
A single model with a standard NMT architecture performs both forward and backward translation, allowing it to back-translate and incorporate any source or target monolingual data.
By continuing training on augmented parallel data, bi-directional NMT models consistently achieve improved translation quality, particularly in low-resource scenarios and cross-domain tasks.
These models also reduce training and deployment costs significantly compared to standard uni-directional models.

\section*{Acknowledgments}

Part of this research was conducted while the first author was an intern at Amazon.
At Maryland, this research is based upon work supported in part by the Clare Boothe Luce Foundation, and by the Office of the Director of National Intelligence (ODNI), Intelligence Advanced Research Projects Activity (IARPA), via contract \#FA8650-17-C-9117. The views and conclusions contained herein are those of the authors and should not be interpreted as necessarily representing the official policies, either expressed or implied, of ODNI, IARPA, or the U.S. Government. The U.S. Government is authorized to reproduce and distribute reprints for governmental purposes notwithstanding any copyright annotation therein.

\bibliography{bi-nmt}

\begin{thebibliography}{35}
\expandafter\ifx\csname natexlab\endcsname\relax\def\natexlab#1{#1}\fi

\bibitem[{Axelrod et~al.(2011)Axelrod, He, and Gao}]{AxelrodHG11}
Amittai Axelrod, Xiaodong He, and Jianfeng Gao. 2011.
\newblock Domain adaptation via pseudo in-domain data selection.
\newblock In \emph{{EMNLP}}, pages 355--362. {ACL}.

\bibitem[{Ba et~al.(2016)Ba, Kiros, and Hinton}]{BaKH16}
Lei~Jimmy Ba, Ryan Kiros, and Geoffrey~E. Hinton. 2016.
\newblock Layer normalization.
\newblock \emph{CoRR}, abs/1607.06450.

\bibitem[{Bahdanau et~al.(2015)Bahdanau, Cho, and Bengio}]{BahdanauCB15}
Dzmitry Bahdanau, Kyunghyun Cho, and Yoshua Bengio. 2015.
\newblock Neural machine translation by jointly learning to align and
  translate.
\newblock In \emph{{ICLR}}.

\bibitem[{Bojar et~al.(2017)Bojar, Chatterjee, Federmann, Graham, Haddow,
  Huang, Huck, Koehn, Liu, Logacheva, Monz, Negri, Post, Rubino, Specia, and
  Turchi}]{BojarCFGHHHKLLM17}
Ondrej Bojar, Rajen Chatterjee, Christian Federmann, Yvette Graham, Barry
  Haddow, Shujian Huang, Matthias Huck, Philipp Koehn, Qun Liu, Varvara
  Logacheva, Christof Monz, Matteo Negri, Matt Post, Raphael Rubino, Lucia
  Specia, and Marco Turchi. 2017.
\newblock Findings of the 2017 conference on machine translation {(WMT17)}.
\newblock In \emph{{WMT}}, pages 169--214. Association for Computational
  Linguistics.

\bibitem[{Brislin(1970)}]{Brislin70}
Richard~W Brislin. 1970.
\newblock Back-translation for cross-cultural research.
\newblock \emph{Journal of cross-cultural psychology}, 1(3):185--216.

\bibitem[{Buck et~al.(2014)Buck, Heafield, and van Ooyen}]{BuckHO14}
Christian Buck, Kenneth Heafield, and Bas van Ooyen. 2014.
\newblock N-gram counts and language models from the common crawl.
\newblock In \emph{{LREC}}, pages 3579--3584. European Language Resources
  Association {(ELRA)}.

\bibitem[{Burton et~al.(2009)Burton, Java, and Soboroff}]{BurtonJS09}
Kevin Burton, Akshay Java, and Ian Soboroff. 2009.
\newblock \href {http://www.icwsm.org/data/} {The {ICWSM} 2009 {Spinn3r}
  dataset}.
\newblock In \emph{Proceedings of the Third Annual Conference on Weblogs and
  Social Media ({ICWSM} 2009), San Jose, CA}.

\bibitem[{Caruana(1997)}]{Caruana97}
Rich Caruana. 1997.
\newblock Multitask learning.
\newblock \emph{Machine Learning}, 28(1):41--75.

\bibitem[{Cettolo et~al.(2017)Cettolo, Federico, Bentivogli, Niehues,
  St{\"u}ker, Sudoh, Yoshino, and Federmann}]{IWSLT2017}
Mauro Cettolo, Marcello Federico, Luisa Bentivogli, Jan Niehues, Sebastian
  St{\"u}ker, Katsuitho Sudoh, Koichiro Yoshino, and Christian Federmann. 2017.
\newblock Overview of the {IWSLT} 2017 evaluation campaign.
\newblock In \emph{International Workshop on Spoken Language Translation},
  pages 2--14.

\bibitem[{Cheng et~al.(2016)Cheng, Xu, He, He, Wu, Sun, and
  Liu}]{ChengXHHWSL16}
Yong Cheng, Wei Xu, Zhongjun He, Wei He, Hua Wu, Maosong Sun, and Yang Liu.
  2016.
\newblock Semi-supervised learning for neural machine translation.
\newblock In \emph{{ACL} {(1)}}. The Association for Computer Linguistics.

\bibitem[{Cho et~al.(2014)Cho, van Merrienboer, G{\"{u}}l{\c{c}}ehre, Bahdanau,
  Bougares, Schwenk, and Bengio}]{ChoMGBBSB14}
Kyunghyun Cho, Bart van Merrienboer, {\c{C}}aglar G{\"{u}}l{\c{c}}ehre, Dzmitry
  Bahdanau, Fethi Bougares, Holger Schwenk, and Yoshua Bengio. 2014.
\newblock Learning phrase representations using {RNN} encoder-decoder for
  statistical machine translation.
\newblock In \emph{{EMNLP}}, pages 1724--1734. {ACL}.

\bibitem[{Christodoulopoulos and Steedman(2015)}]{Christodoulopoulos15}
Christos Christodoulopoulos and Mark Steedman. 2015.
\newblock A massively parallel corpus: the bible in 100 languages.
\newblock \emph{Language Resources and Evaluation}, 49(2):375--395.

\bibitem[{Chu et~al.(2017)Chu, Dabre, and Kurohashi}]{ChuDK17}
Chenhui Chu, Raj Dabre, and Sadao Kurohashi. 2017.
\newblock An empirical comparison of domain adaptation methods for neural
  machine translation.
\newblock In \emph{{ACL} {(2)}}, pages 385--391. Association for Computational
  Linguistics.

\bibitem[{Denil et~al.(2013)Denil, Shakibi, Dinh, Ranzato, and
  de~Freitas}]{DenilSDRF13}
Misha Denil, Babak Shakibi, Laurent Dinh, Marc'Aurelio Ranzato, and Nando
  de~Freitas. 2013.
\newblock Predicting parameters in deep learning.
\newblock In \emph{{NIPS}}, pages 2148--2156.

\bibitem[{Dong et~al.(2015)Dong, Wu, He, Yu, and Wang}]{DongWHYW15}
Daxiang Dong, Hua Wu, Wei He, Dianhai Yu, and Haifeng Wang. 2015.
\newblock Multi-task learning for multiple language translation.
\newblock In \emph{{ACL} {(1)}}, pages 1723--1732. The Association for Computer
  Linguistics.

\bibitem[{Firat et~al.(2016)Firat, Cho, and Bengio}]{FiratCB16}
Orhan Firat, Kyunghyun Cho, and Yoshua Bengio. 2016.
\newblock Multi-way, multilingual neural machine translation with a shared
  attention mechanism.
\newblock In \emph{{HLT-NAACL}}, pages 866--875. The Association for
  Computational Linguistics.

\bibitem[{G{\"{u}}l{\c{c}}ehre et~al.(2015)G{\"{u}}l{\c{c}}ehre, Firat, Xu,
  Cho, Barrault, Lin, Bougares, Schwenk, and Bengio}]{GulcehreFXCBLBS15}
{\c{C}}aglar G{\"{u}}l{\c{c}}ehre, Orhan Firat, Kelvin Xu, Kyunghyun Cho,
  Lo{\"{\i}}c Barrault, Huei{-}Chi Lin, Fethi Bougares, Holger Schwenk, and
  Yoshua Bengio. 2015.
\newblock On using monolingual corpora in neural machine translation.
\newblock \emph{CoRR}, abs/1503.03535.

\bibitem[{He et~al.(2016)He, Xia, Qin, Wang, Yu, Liu, and Ma}]{HeXQWYLM16}
Di~He, Yingce Xia, Tao Qin, Liwei Wang, Nenghai Yu, Tie{-}Yan Liu, and
  Wei{-}Ying Ma. 2016.
\newblock Dual learning for machine translation.
\newblock In \emph{{NIPS}}, pages 820--828.

\bibitem[{Hieber et~al.(2017)Hieber, Domhan, Denkowski, Vilar, Sokolov,
  Clifton, and Post}]{CoRR:Sockeye}
Felix Hieber, Tobias Domhan, Michael Denkowski, David Vilar, Artem Sokolov, Ann
  Clifton, and Matt Post. 2017.
\newblock Sockeye: {A} toolkit for neural machine translation.
\newblock \emph{CoRR}, abs/1712.05690.

\bibitem[{Johnson et~al.(2017)Johnson, Schuster, Le, Krikun, Wu, Chen, Thorat,
  Vi{\'{e}}gas, Wattenberg, Corrado, Hughes, and Dean}]{JohnsonSLKWCTVW17}
Melvin Johnson, Mike Schuster, Quoc~V. Le, Maxim Krikun, Yonghui Wu, Zhifeng
  Chen, Nikhil Thorat, Fernanda~B. Vi{\'{e}}gas, Martin Wattenberg, Greg
  Corrado, Macduff Hughes, and Jeffrey Dean. 2017.
\newblock Google's multilingual neural machine translation system: Enabling
  zero-shot translation.
\newblock \emph{{TACL}}, 5:339--351.

\bibitem[{Junczys{-}Dowmunt et~al.(2016)Junczys{-}Dowmunt, Dwojak, and
  Sennrich}]{Junczys-Dowmunt16}
Marcin Junczys{-}Dowmunt, Tomasz Dwojak, and Rico Sennrich. 2016.
\newblock The {AMU-UEDIN} submission to the {WMT16} news translation task:
  Attention-based {NMT} models as feature functions in phrase-based {SMT}.
\newblock In \emph{{WMT}}, pages 319--325. The Association for Computer
  Linguistics.

\bibitem[{Kingma and Ba(2015)}]{KingmaB15}
Diederik~P. Kingma and Jimmy Ba. 2015.
\newblock Adam: {A} method for stochastic optimization.
\newblock In \emph{{ICLR}}.

\bibitem[{Koehn and Knowles(2017)}]{KoehnK17}
Philipp Koehn and Rebecca Knowles. 2017.
\newblock Six challenges for neural machine translation.
\newblock In \emph{NMT@ACL}, pages 28--39. Association for Computational
  Linguistics.

\bibitem[{Koehn et~al.(2003)Koehn, Och, and Marcu}]{KoehnOM03}
Philipp Koehn, Franz~Josef Och, and Daniel Marcu. 2003.
\newblock Statistical phrase-based translation.
\newblock In \emph{{HLT-NAACL}}. The Association for Computational Linguistics.

\bibitem[{Koehn and Schroeder(2007)}]{KoehnS07}
Philipp Koehn and Josh Schroeder. 2007.
\newblock Experiments in domain adaptation for statistical machine translation.
\newblock In \emph{WMT@ACL}, pages 224--227. Association for Computational
  Linguistics.

\bibitem[{Moore and Lewis(2010)}]{MooreL10}
Robert~C. Moore and William~D. Lewis. 2010.
\newblock Intelligent selection of language model training data.
\newblock In \emph{{ACL} (Short Papers)}, pages 220--224. The Association for
  Computer Linguistics.

\bibitem[{Nguyen and Chiang(2018)}]{NguyenC18}
Toan~Q. Nguyen and David Chiang. 2018.
\newblock Improving lexical choice in neural machine translation.
\newblock In \emph{{HLT-NAACL}}. The Association for Computational Linguistics.

\bibitem[{Press and Wolf(2017)}]{PressW17}
Ofir Press and Lior Wolf. 2017.
\newblock Using the output embedding to improve language models.
\newblock In \emph{{EACL} {(2)}}, pages 157--163. Association for Computational
  Linguistics.

\bibitem[{Ramachandran et~al.(2017)Ramachandran, Liu, and
  Le}]{RamachandranLL17}
Prajit Ramachandran, Peter~J. Liu, and Quoc~V. Le. 2017.
\newblock Unsupervised pretraining for sequence to sequence learning.
\newblock In \emph{{EMNLP}}, pages 383--391. Association for Computational
  Linguistics.

\bibitem[{Sennrich et~al.(2016{\natexlab{a}})Sennrich, Haddow, and
  Birch}]{SennrichHB16}
Rico Sennrich, Barry Haddow, and Alexandra Birch. 2016{\natexlab{a}}.
\newblock Improving neural machine translation models with monolingual data.
\newblock In \emph{{ACL} {(1)}}. The Association for Computer Linguistics.

\bibitem[{Sennrich et~al.(2016{\natexlab{b}})Sennrich, Haddow, and
  Birch}]{SennrichHB16a}
Rico Sennrich, Barry Haddow, and Alexandra Birch. 2016{\natexlab{b}}.
\newblock Neural machine translation of rare words with subword units.
\newblock In \emph{{ACL} {(1)}}. The Association for Computer Linguistics.

\bibitem[{Sutskever et~al.(2014)Sutskever, Vinyals, and Le}]{SutskeverVL14}
Ilya Sutskever, Oriol Vinyals, and Quoc~V. Le. 2014.
\newblock Sequence to sequence learning with neural networks.
\newblock In \emph{{NIPS}}, pages 3104--3112.

\bibitem[{Wang et~al.(2018)Wang, Xia, Zhao, Bian, Qin, Liu, and
  Liu}]{WangXZBQLL18}
Yijun Wang, Yingce Xia, Li~Zhao, Jiang Bian, Tao Qin, Guiquan Liu, and
  Tie{-}Yan Liu. 2018.
\newblock Dual transfer learning for neural machine translation with marginal
  distribution regularization.
\newblock In \emph{{AAAI}}. {AAAI} Press.

\bibitem[{Zhang and Zong(2016)}]{ZhangZ16}
Jiajun Zhang and Chengqing Zong. 2016.
\newblock Exploiting source-side monolingual data in neural machine
  translation.
\newblock In \emph{{EMNLP}}, pages 1535--1545. The Association for
  Computational Linguistics.

\bibitem[{Zhang et~al.(2018)Zhang, Liu, Li, Zhou, and Chen}]{ZhangLLZC18}
Zhirui Zhang, Shujie Liu, Mu~Li, Ming Zhou, and Enhong Chen. 2018.
\newblock Joint training for neural machine translation models with monolingual
  data.
\newblock In \emph{{AAAI}}. {AAAI} Press.

\end{thebibliography}
\bibliographystyle{acl_natbib}

\end{document}